\documentclass{article} 
\usepackage{aaai,times}
\usepackage{hyperref}
\usepackage{url}

\usepackage{graphicx}
\usepackage{amsmath}

\DeclareGraphicsExtensions{.jpg, .pdf, .png}

\title{Towards A Model Theory for Distributed Representations}

\author{
R.V.Guha \\
  Google \\
  \texttt{guha@guha.com}\\
}

\begin{document}

\maketitle

\begin{abstract}

 Distributed representations (such as those based on embeddings) and
symbolic representations (such as those based on logic) have complementary
strengths. We explore one possible approach to combining these two
kinds of representations. We present
a model theory/semantics for first order logic based on vectors of reals. We 
describe the model theory and discuss some interesting properties of such
a representation.

\end{abstract}

 \section{Introduction}

 Knowledge Representation based approaches to AI involve 
encoding knowledge in a logical language and performing logical
inference to derive conclusions. Such systems have 
certain highly desirable properties.
\begin{itemize}
\item They are {\bf teachable}. We can add both specific facts and general 
axioms/heuristics concisely. E.g.,
we can simply tell such a system that every human has a biological mother, without
having to feed it a large number of examples, in the hope that a learning
system appropriately generalizes
\item There is a well defined notion of {\bf entailment}, that
allows us to draw conclusions from the general axioms we add to the system
\end{itemize}

 These systems, which are usually based on some form of first order
 logic, are very good for writing axioms to represent (and reason
 about) complex domains.
These axioms are typically hand written, because of which building a broad artificial
intelligence using this approach has proven to be rather daunting \cite{cyc}. 
 Completely automating the construction of these systems using learning has also
 proven difficult. Complex first order statements are extremely hard to
 automatically learn.

The strengths of Knowledge Representation based system come from the origins of these systems, namely, in 
mathematical logic. Unfortunately, these origins also bring some unwanted
baggage. Mathematical logic was developed for the purpose of stating mathematical
truths in a manner where the terms in these statements have
precise and unambiguous meaning.
Most axioms we
add to our knowledge based systems are transliterations of natural language 
utterances. And as with all such utterances, despite our best attempts, terms
and axioms in knowledge based systems end up having many of the characteristics of 
natural language. In particular, our experiences with systems such as Cyc \cite{cyc} and
Schema.org \cite{schema} have highlighted the fluid and ambiguous
nature of linguistic terms. Not just concepts like 'chair', but even terms like 'person' 
afford a wide range of meanings, something difficult for logic based
systems to handle.

Recent work on distributed representations [\cite{socher}, \cite{manning}, 
\cite{bordes}, \cite{bordes2014semantic}, \cite{quoc}] has explored the use of embeddings as a representation
tool. These approaches typically 'learn an embedding', which maps terms and statements
in a knowledge base (such as Freebase \cite{freebase}) to points in an N-dimensional vector space.
Vectors between points can then be interpreted as relations between
the terms. A very attractive property of these distributed
representations is the fact that they are learnt from a set of
examples. Further, the continuous nature of the underlying vector space also
gives hope for coping with the fluidity encountered in the meaning of terms.

 But this benefit comes at the cost of not being able to do some of the
things that are relatively trivial for logic based systems. 

\section{Goals \& Outline of Approach}
We would like to have systems that are largely learnt, which we can also teach. 
In this work we take the first steps towards building a representation system 
that combines the strengths of logical and distributed representations. The first step  is
to create a system that has a common representation for both
embeddings and logical sentences. The representation needs to be
common not just in syntax, but also in terms of semantics, i.e., in
what operations can be carried out on them.

Model theory \cite{enderton} is the mathematical foundation for logic. It 
tells us what logical sentences may be construed to mean, which
operations make sense and what can be said to follow from a set of
statements in a knowledge base. We believe that an essential step in
bringing logic and distributed representations closer is to create a
model theory based on embeddings.

Rather than look at the geometric properties of learnt 
embeddings as validation of the system having a semantic understanding, we take 
the concept of an embedding as the starting point and try to build a model theory out of
it.

Our model theory is structurally similar to the standard
Tarskian semantics for first order logic. Tarskian semantics  is based on
the concept of an interpretation for a set of statements in a
language. An interpretation maps symbols in the language 
into objects and relations (n-tuples of these objects).
In contrast, our interpretations map symbols in the
language to points and vectors in an
N-dimensional space. 
Intuitively, a good/correct embedding maps to a single satisfying
interpretation. We define satisfaction and entailment as in Tarskian
semantics. 

This small change (from objects to points) in Tarskian semantics is not enough to reflect
object similarity as captured by the geometry of embeddings. To
recapture this, we introduce a class of preferred models, where the
relative geometric location of objects reflects their similarity. 
We argue that such models, where similar objects are spatially closer,
better capture the generalizations implicit in the data. 
We present an approach to simple inference in these preferred
models. 

Finally we revisit some old thorny
problems that come up in representing common sense knowledge and
discuss how a vector space approach might help.

This paper is an early exploration along this direction.
Much work needs to be done before we can actually build systems based
on the approaches described here.

\section{Model Theory}

\subsection{Recap of Tarskian Semantics}
For the sake of simplicity, without loss of generality, we restrict our attention to logical languages
with no function symbols, no free variables and with only binary predicates.

Tarskian semantics for first order logic is based on
the concept of an interpretation for a set of logical statements in a
language. The interpretation is defined using a model. 
A model for a first order language assigns an interpretation to all 
the non-logical constants in that language. More specifically,

\begin{enumerate}
\item A model  $M$ specifies a set of objects $D$ ($d_1$,
  $d_2$, ...), the domain of discourse.

\item To each term $t_i$ in the language, $M$ assigns an object in
  $M(t_i)$ in $D$

\item Each (binary) predicate symbol $P$ is assigned to a relation $M(P)$
  over $D^2$
\end{enumerate}

A sentence in the language evaluates to $True$ or $False$ given a
model  $M$ if
\begin{enumerate}
\item Atomic formulas: A formula $P(t_1, t_2)$ evaluates to $True$ iff
  $<d_{t1}, d_{t2}>  \in  M(P)$
\item Formulas with logical connectives, such as $\neg \phi$, $\phi
  \rightarrow \psi$ are evaluated according to propositional truth tables
\item $\exists x \phi(x)$ is $True$ if there exists some element of
  $D$, $d_i$ for which $\phi(d_i)$ is true.
\item  $\forall x \phi(x)$ is true if $\phi(d_i)$ is true for every
  element  $ d_i \in D$.
\end{enumerate}

If a sentence $\phi$ evaluates to $True$ under a given interpretation $M$, one
says that M satisfied $\phi$; this is denoted $M \models \phi$. A sentence is
satisfiable if there is some interpretation/model under which it is $True$. A 
formula is logically valid (or simply valid) if it is $True$ in every
interpretation. 

A formula $\psi$ is a logical consequence of a formula $\phi$ if every
interpretation that makes $\phi$ $True$ also makes $\psi$ $True$. 
In this case one says that $\psi$ is logically implied by $\phi$. 
It is this notion of logical implication that allows us to do
inference in knowledge based systems.

\subsection{Embeddings based Semantics}

We now describe how Tarskian semantics can be modified to be based on
a vector space model. We do this by using a different
kind of model, wherein the domain is a set of points in an
$N$ dimensional vector space of reals. More specifically,

\begin{enumerate}
\item A model  $M$ specifies an $N$ dimensional vector space.

\item To each term $t_i$ in the language, $M$ assigns a point 
  $M(t_i)$ in this vector space

\item Each (binary) predicate symbol $P$ is assigned to a unit vector $M(P)$ in
  $K \le N$ dimensions of the vector space.\footnote{If $M(P)$ is $N$
    dimensions, then if $P(A, B)$ and $P(A, C)$, $B$ will have to be
    equal to $C$. Hence $M(P)$ is in a subspace.} 

\end{enumerate}

$P(t_1, t_2)$  evaluates to $True$ iff the projection of the vector
from $M(t_1)$ to $M(t_2)$  onto the K dimensions of $M(P)$ 
has the same direction as $M(P)$.\\

The definitions for evaluating formulas with logical
constants, formulas with quantifiers, of satisfaction and logical
entailment  are the same as with Tarskian semantics.

Each of our models is also a Tarskian model in a fashion that is
consistent with the Tarskian definition of satisfaction, entailment,
etc. Consequently, the soundness of classical inference rules (modus
ponens, resolution, etc.) carry over.

This kind of model corresponds closely to the kind of embeddings (TransE)
described in \cite{bordes2014semantic}. In that work, the authors present a
mechanism for computing mappings from terms to points in the 
vector space that is maximally consistent with and predictive of a
database of atomic formulas such as those in
Freebase. \cite{wang2014knowledge} solve the problem of representing
one to many relations in their TransH model.
In \cite{quoc}
the authors use a similar approach to map words (Paris, France, Rome,
Italy, etc.) into a vector space so as to maximize skipgram recall. The
vectors between pairs such as (Paris, France) tend out to be parallel
to those between (Rome, Italy), i.e., are 'semantically' meaningful.

We have taken this as a starting point, but instead of treating such
embeddings, where terms/words map to points in an N-dimensional
vector space and relations map to vectors, as the target of a 
learning function, we have used them as the starting point for a model
theory.

\section{Spatial aspects of models}

  We have simply mapped objects to points in a real space. By itself,
this does not solve any of the issues faced by purely symbol
representations.  The real benefits can come only when we exploit the 
fact that this is in a 'space' and associate
meaning with the absolute/relative locations of objects beyond
associating relations with vectors.

\subsection{Aggregate/Approximate models}

Certain concepts (e.g., the number one) will have a fairly crisp
meaning, whereas certain other concepts (e.g., chair), can have a
rather broad/vague/approximate meaning. Systems based on
logic have found it very difficult to capture this. We introduce
Aggregate and Approximate models, two alternatives that both use continuous nature of
the embedding space offers the hope of being able to capture this.

\subsubsection{Approximate Models} The simplest way to incorporate 
approximateness into our model is to allow some variation in the vector
corresponding to each relation, which in turn allows for some variation 
in the location of of each object.

\subsubsection{Aggregate Models} Consider the set of points across different
models corresponding to a particular term. Consider a cluster
of these points (from a subset of the satisfying models)
which are sufficiently close to each other. This cluster or cloud 
of points (each of which is in a different model), corresponds to $an$ aggregate
of possible interpretations of the term. We can extend this approach for 
all the (atomic) terms in the language. We
pick a subset of models where every term forms such a cluster. The set of clusters
and the vectors between  them gives us the aggregate model. Note that
in vectors corresponding to relations will also allow amount of variation.
If a model satisfies the KB, any linear transform of the model will
also satisfy the KB. In order to keep these transforms from taking
over, no two models that form an aggregate should be linear transforms
of each other.

In both aggregate and approximate models, each object corresponds to a cloud in the 
N-dimensional space and the relation between objects is captured by their approximate
relative positions. The size of the cloud corresponds to the 
vagueness/approximateness (i.e., range of possible meanings) of the concept.

\subsection{Learning and object locations}

Learnt embeddings, such as those reported in \cite{wang2014knowledge} and
\cite{bordes2014semantic} have the property that similar objects
tend to be spatially closer to each other than to objects that are
disimilar. This is a result of the learning/optimization mechanisms by which embeddings
are created. Given a KB of ground atomic facts (triples), these
systems are trained  on a subset of these triples. The output of the
training is a set of locations for the objects.
The goal of the learning is to correctly predict the other triples in
the KB. In other words, even though the KB itself does not contain
general axioms, the learning discovers the implicit, general axioms
governing the facts of the domain. These general axioms are not
directly represented in the learnt embedding, but are reflected in the
placement of the objects in the vector space. 

We now try to explain why embeddings where similar objects are closer
tend to capture generalizations in the domain and should hence be
preferred. Imagine that there are a set of axioms that capture the
generalities in the domain. They imply some subset of the
triples in the KB from the other triples. The goal of the learning
algorithm is to 'discover' the implications of these axioms. 

We make our case on a class of axioms that is simple, but very important. 
Consider axioms of the form \\

$(\forall \; x \; P(x, A) \implies Q(x, B))$ \\

where P, Q are predicates and A, B are constants. Though this axiom
template looks very simple, in systems like \cite{cyc}, a significant
fraction of axioms follow this pattern. Of note are inheritance rules, which have the form \\

$(\forall \; x \; isa(x, \langle Category \rangle) \implies Q(x, \langle Attribute \rangle)$ \\

In our model, P and Q map
to vectors  $P_v$ and $Q_v$ in some subspace of the
N-dimensional space. Given two objects $x_1$ and $x_2$ such that
$P(x_1, A)$ and $P(x_2, A)$, $x_1$ and $x_2$ will share the same
coordinates in the subspace of $P_v$ and differ in their coordinates
in the other dimensions. It is easy to see that the likelihood of
$Q(x_1, B)$ and $Q(x_2, B)$ also being true (in the learnt embedding)
is higher if $x_1$ and $x_2$ are close in these other dimensions as
well. In other words, if the learning system is given a some of triples
of the form $P(x_i, A)$ and some of the form $Q(x_i, B)$, where there
is an overlap in the $x_i$, by placing these $x_i$, which share the
similarity that $P(x_i, A)$ is true of them, close together, it
increases the likelihood of correctly predicting $Q(x_i, B)$. 

Applying this observation to inheritance rules, since objects typically
inherit some properties by virtue of what kind of object they are, 
it follows that objects of the same type are likely to be found close to each
other in the embedding space.

In other words, of the set all satisfying models, the subset of models
in which objects of the same type (or more generally, similar objects) are 
placed together, better capture the generalities implicit in the data.

Coming back to our model theory, unfortunately, though we map 
terms to points in our models,  there is
no similarity metric that is built into this mapping. Consequently,
even in satisfying models, points that are
extremely close  may denote extremely dissimilar terms.
Further, in order to determine if something is logically entailed by a knowledge base, 
we have to consider the set of all models that satisfy the knowledge
base. Different satisfying models might have completely
different coordinates for the same term and different vectors for
the same predicate. 

We now introduce a class of preferred models which
try to capture this intuition.

\subsection{Preferred Models}

 Typically, in machine learning based approaches, a number of
examples are used  to try construct a single model or a probability
distribution over models. There is a tradeoff between precision and
recall, where we tolerate some number of wrong predictions in order to
increase the number of correct predictions.

Logical approaches on the other hand, try to
get $all$ the correct predictions (i.e., completeness) while avoiding
$all$ wrong predictions (i.e., soundness). To do this they deal not
with a  single 'best' model, but with the set of all satisfying
models. The only statements that follow are those that are true in
$all$ these models.
For example, consider a knowledge base about American history.
It will likely contain a symbol like 'AbrahamLincoln', which the author of the KB
$intends$ to denote the 16th American President. The logical
machinery doesn't care if the satisfying models map it to the
President, flying cars, real numbers or just the symbol itself. It
will draw a conclusion $only$ if the conclusion follows under $all$
these interpretations that satisfy the KB.
This is at the heart of soundness in logical inference.


Research in non-monotonic reasoning has explored relaxing the 
heavy constraint of only drawing conclusions true in
all satisfying models. For example, circumscription
[\cite{hintikka}, \cite{circumscription}]
allows conclusions that are true in only a 
preferred subset of satisfying models, those that minimize
the extent of certain predicates (typically the 'ab' predicates).
Such systems sacrifice soundness for the sake of non-monotonicity.

We follow a similar path, introducing a class of preferred models that
sacrifice soundness for the sake of learning generalizations implicit
in the data. We made the case earlier that models where object similarity
is proportional to object distance better capture generalities in
the domain. 

We use similarity to define a set of preferred models. 
Assume that we have a similarity
function $S(t_1, t_2)$ which measures the similarity between two
terms and evaluates to a number between 0 and 1, with $S(t_1, t_2)$
being closer to 1 if $t_1$ and $t_2$ are more similar. We want models
where the distance between the points denoting $t_1$ and $t_2$ is
correlated (inversely) to $S(t_1, t_2)$. When this is the case for
every pair of points, we have model where the geometry has
significance. Let $D(t_i, t_j)$ be the distance
between the points that $t_i$ and $t_j$ map to. Then when \\

$SD(t_i, t_j) = (1 - S(t_i, t_j))/D(t_i,t_j) \approx 1$ \\

for every pair of terms, the proximity in the model correlates with
similarity between the objects denoted by the terms. There are 
multiple ways of picking such models. For example, we can 
minimize  \\

$(\Sigma_{i=0}^L \Sigma_{j=0}^L log(SD(t_i, t_j)))/ L^2$ \\

where L is the number of terms This measures the average disparity between the
similarity measure and the distance between (dis)similar objects.
The preferred models are those where this average is less than some
threshold. Alternately, we can pick all models where a measure such as
$log(SD(t_i, t_j))$ is less than some threshold for each
pair of terms.

\subsection{Inference and Learning}
As mentioned earlier, since every model in our framework is also
Tarskian model and our definition of satisfaction and entailment are
the same, every sound proof procedure for first order logic is also
sound in our system. 

However, we would like inference mechanisms that are cognizant of
preferred and aggregate models, i.e., that exploit the geometric structure of our
models. Much work needs to be done before we have
general  purpose inference mechanisms such as resolution, 
that exploit the geometric properties of preferred models.

One approach to approximate inference, that works for domains with a small
number of objects, is as follows. We build a representative ensemble
of preferred approximate models.
Queries are answered by checking the query against each of the
models in the ensemble. If the query formula is true (or is true for
the same variable bindings) in every model in the ensemble, it is true.
Since the model checking is only done over preferred models, 
the result should be a combination of learning and logical inference.

\subsection{Model Generation}

Here, we present a  simple approach for 
generating a set of preferred models that are consistent with a given KB.
 We map axioms (ground and quantified) in the KB to equations/constraints on 
the coordinates of objects. Solutions to these constraints correspond to 
our preferred models. This approach works only for KBs with small enumerable
domains.

\subsubsection{Ground Atomic Formulas} 
Each triple $P_i(t_j, t_k)$ gives us the equation: \\

$|M(t_j, P_i) - M(t_k, P_i) - M(P_i)| < \delta$ \\

where $M(t_j, P_i)$ is the location in the $K$ dimensional subspace
corresponding to $P_i$ of the term
$t_j$ and $M(P_i)$ is the vector corresponding to $P_i$ and $\delta$
is some measure of the approximateness we want to allow in our relations.

\subsubsection{Quantified Axioms} 
Next, we need a set of equations that capture the quantified
axioms. We 'unroll' the quantifiers by instantiatinting the quantified variables
with the terms in the KB and then map the instantiated ground axioms
to equations.

We illustrate our approach on a simple class of axioms. 
Consider axioms of the form $(\forall \; x \; P(x,A) \implies Q(x, B))$.
We instantiate this axiom for each term $t_i$. Consider a ground
non-atomic formula such as \\

$P(t_i, A) \implies Q(t_i, B) \equiv \neg P(t_i, A) \vee Q(t_i, B)$ \\

We map $Q(t_i, B)$ to 
$\sigma (M(t_i, Q_i) - M(B, Q_i) - M(Q))$ \\
 where $\sigma$ is a sigmoid function that is 1 if $|M(t_i, Q) - M(B, Q) -
M(Q))| < \delta$ and 0 otherwise. \\

$\neg P(t_i, A)$ maps to
$(\sigma^{-1} (M(t_i, P) - M(A, P) - M(Q)))$ \\

Disjunctions are modelled by addition. So, \\
$P(t_i, A) \implies Q(t_i, B)$ is mapped to \\

$\sigma^{-1} (M(t_i, P) - M(A, P) - M(P)) + \\
\; \; \;\sigma (M(t_i, Q) - M(B, Q) - M(Q)) = 1$ \\

More complex formulas can similarly be used to generate more
constraints on the locations of the objects. 

Finally, we have a set of  constraints based on the similarity 
function $S(t_j, t_k)$, which try to learn generalities implicit
in the data by placing similar objects close to each other.

A variety of existing techniques can be used to solve this system
of constraints.

\section{Further thoughts}

The vector space representation very loosely corresponds to a
’semantics’  or ’understanding’ that the system has. In logical 
systems, the semantics is only in the meta-theory of the 
system (i.e., governs what the system should and should 
not do), not in the system itself.

Having a set of structures, distinct from the logical
  statements, that correspond to the system's understanding gives us a
  mechanism for dealing with the variation and context sensitivity
  in the meaning of terms. The same term, in different statements
  could map to slightly different points in the vector space, thereby
  having slightly different meanings.

A vector space based model gives us a generative function for
objects. Consider a symbol in the language (e.g., 'Chair'). In
classical semantics, this symbol denotes a single object in a given
model. There $may$ be other objects in the model that are very
similar, but lacking a term that refers to them. The discreteness of
the Tarskian model puts beyond the reach of our language. Attempts
to incorporate context into logic (\cite{guha_contexts},
\cite{jmc_contexts} allow for different occurances of a term to
refer to distinct objects, but do so at the relatively heavy cost of
making them completely different. We are hopeful that the vector space model 
might give us a tool that gives us a more nuanced control over the
denotation space. We feel that this is one of the biggest promises of
this approach.

\section{Conclusions}

 In this paper, we took some first steps towards building a
 representation system that combines the benefits of traditional logic
 based systems and systems based on distributed representations.
We sketched the outline of a Model Theory for  a logic, along the 
lines of Tarskian semantics, but based on vector spaces. We introduced
a class of preferred models that capture the geometric intuitions
behind vector space models and outlined a model checking based
approach to answering simple queries.
 
\section{Acknowledgements}
 I would like to thank Vineet Gupta, Evgeniy Gabrilovich, Kevin
 Murphy, John Giannandrea
and Yoram Singer for feedback and discussions.

\bibliographystyle{aaai}
\bibliography{emt}

\end{document}